\documentclass[letterpaper, 10 pt, conference]{ieeeconf}  
\usepackage[utf8]{inputenc}
\usepackage[T1]{fontenc}

\IEEEoverridecommandlockouts                              

\usepackage[pdftex]{graphicx} 
\usepackage{amsfonts}
\usepackage{amssymb}  
\usepackage{bm}
\usepackage{booktabs}
\usepackage{hyperref}
\usepackage{microtype}
\usepackage[super]{nth}
\usepackage{siunitx}
\usepackage{stmaryrd}
\usepackage[flushleft]{threeparttable}

\usepackage[fixamsmath]{mathtools}
\DeclarePairedDelimiter\parentheses{\lparen}{\rparen}

\newcommand{\func}[2]{\ensuremath{#1 \parentheses*{ #2 }}}

\usepackage{silence}
\usepackage[backend=biber, style=ieee, maxnames=6, minnames=3, natbib=true]{biblatex}
\AtBeginBibliography{\small}
\usepackage{cleveref}
\crefname{figure}{Fig.}{Figs.}
\Crefname{figure}{Fig.}{Figs.}
\crefname{table}{Table}{Tables}
\Crefname{table}{Table}{Tables}
\crefname{equation}{Eq.}{Eqs.}
\crefname{equation}{Eq.}{Eqs.}

\newcommand{\train}{\textsc{train}}
\newcommand{\eval}{\textsc{eval}}
\newcommand{\test}{\textsc{test}}

\newcommand{\mbl}{\mathcal{M}_{\mathrm{FM}}}
\newcommand{\mpt}{\mathcal{M}_{\mathrm{PT}}}
\newcommand{\mft}{\mathcal{M}_{\mathrm{FT}}}

\newcommand{\real}{\mathbb{R}}
\newcommand{\N}{\mathbb{N}}

\title{\LARGE \bf
Deep transfer learning for improving single-EEG arousal detection
}

\author{Alexander Neergaard Olesen$^{\ast,1,2,3}$,~\IEEEmembership{Member, IEEE}, Poul Jennum$^{\dagger,3}$, \\Emmanuel Mignot$^{\dagger,2}$ and Helge B. D. Sorensen$^{\dagger,1}$,~\IEEEmembership{Senior Member, IEEE}
\thanks{Research supported by the Klarman Family Foundation, Technical University of Denmark, and University of Copenhagen with supporting grants from Reinholdt W. Jorck \& Wife's Foundation, Knud Højgaard Foundation, Otto Mønsted Foundation, Vera \& Carl Michaelsens Foundation, Augustinus Foundation, and Stibo Foundation.}
\thanks{$^{\ast}$Corresponding author: {\tt aneol@dtu.dk}}%
\thanks{$^{\dagger}$These authors contributed equally.}
\thanks{$^{1}$Department of Health Technology, Technical University of Denmark, 2800 Kgs. Lyngby, Denmark.}%
\thanks{$^{2}$Center for Sleep Sciences and Medicine, Stanford University, Palo Alto, CA 94304, USA.}%
\thanks{$^{3}$Danish Center for Sleep Medicine, University Hospital Copenhagen, 2600 Glostrup, Denmark}%
}

\begin{document}

\maketitle
\thispagestyle{empty}
\pagestyle{empty}

\begin{abstract}
Datasets in sleep science present challenges for machine learning algorithms due to differences in recording setups across clinics.
We investigate two deep transfer learning strategies for overcoming the channel mismatch problem for cases where two datasets do not contain exactly the same setup leading to degraded performance in single-EEG models. 
Specifically, we train a baseline model on multivariate polysomnography data and subsequently replace the first two layers to prepare the architecture for single-channel electroencephalography data.
Using a fine-tuning strategy, our model yields similar performance to the baseline model (F1=0.682 and F1=0.694, respectively), and was significantly better than a comparable single-channel model.
Our results are promising for researchers working with small databases who wish to use deep learning models pre-trained on larger databases.

%
\end{abstract}


\section{INTRODUCTION}


A principal tool in the analysis of sleep is the polysomnography (PSG).
Standard PSGs contain electroencephalography (EEG), electrooculography (EOG), electromyography (EMG) from below the chin and lower limbs, electrocardiography, respiratory effort, and blood oxygenation, which is manually analysed by sleep experts according to guidelines published by the American Academy of Sleep Medicine~\cite{Berry2020}.

Experts score sleep stages and annotate discrete events, such as arousals (short awakenings during sleep, $\leq \SI{15}{\second}$), limb movements, and decreased respiratory effort characterized by apneas (complete cessation of breathing), hypopneas (partial cessation of breathing), and desaturations (decreases in oxygen desaturation).
Low inter-rater reliability has been reported for sleep stages scoring in multiple studies~\cite{Norman2000, Rosenberg2013, Younes2016}, arousals~\cite{Bonnet2007}, and respiratory events~\cite{Magalang2013, Rosenberg2014a}, prompting extensive research in automated methods for sleep analysis~\cite{Olesen2018c, Stephansen2018, Chambon2018, Biswal2018, Phan2018, Carvelli2020, Brink-Kjaer2020}.

Designing reliable and robust systems for automated sleep analysis based on machine learning algorithms often require multiple heterogenous data sources of sufficient size.
However, due to differences in clinical practice, very few datasets in sleep science are standardized with regards to recording setups despite guidelines from the AASM.

In these cases, we end up with a \textit{channel mismatch problem}, in which the overlap between our source and target domains is small, and the domains are possibly disjointed. 
Recent studies have investigated the use of deep transfer learning to solve the channel mismatch problem when training and testing sleep stage classification models~\cite{Phan2019, Phan2019c}.
The authors found that using a fine-tuning strategy significantly improved the performance of sleep stage scoring models when trained on various combinations of EEG and EOG channels.

We present results on using deep transfer learning to address the channel mismatch problem, when the source and target domains differ both in the number and type of channel modalities.
Specifically, our source domain consists of multivariate PSG data comprising left and right central EEG, left and right EOG, and submental EMG recordings, while our target domain consists of only a single central EEG channel.
We show that by employing a simple fine-tuning strategy on a pre-trained network stripped of the initial two layers, we can effectively reach the same level of F1 score as when using the full set of PSG data.

\section{METHODS}
\begin{table*}[tb]
    \centering
    \begin{threeparttable}
        \caption{Network architecture overview.} 
        \label{tab:architecture}
        \begin{tabular}{@{}llllllll@{}}
            \toprule
            \textbf{Module} & \textbf{Layer type} & \textbf{Kernel} & \textbf{Stride} & \textbf{Feature maps} & \textbf{Input size} & \textbf{Output size} & \textbf{Activation} \\ \midrule
            \(\mathbf{x}\) & Input & --- & --- & --- & \( C \times T \) & \( 1 \times C \times T \) & --- \\
            \( \phi_{\mathrm{mix}} \) & 2D convolution & \( (C, 1) \) & \( (1, 1) \) & \( C \) & \( 1 \times C \times T \) & \( C \times 1 \times T \) & ReLU \\ \midrule
            \( \varphi_{\mathrm{conv},1} \) & 2D convolution & \( (1, c) \) & \( (1, s) \) & \( 2f_0 \) & \( C \times 1 \times T \) & \( 2f_0 \times 1 \times T/s \) & --- \\
            & Batch norm. & --- & --- & \(2f_0\) & \(2f_0 \times 1 \times T/s\) & \(2f_0 \times 1 \times T/s\) & ReLU \\
            \( \varphi_{\mathrm{conv},k} \) & 2D convolution & \( (1, c) \) & \( (1, s) \) & \( f_02^{k} \) & \( f_02^{k-1} \times 1 \times T/s^{k-1} \) & \( f_02^{k} \times 1 \times T/s^k \) & --- \\
            \( k \in \llbracket 2, k_{\mathrm{max}} \rrbracket \) & Batch norm. & --- & --- & \(f_02^{k}\) & \(f_02^{k} \times 1 \times T/s^k\) & \(f_02^{k} \times 1 \times T/s^k\) & ReLU \\
            \( \varphi_{\mathrm{rec}} \) & bGRU & --- & --- & $f^{\prime}$ & \( f^{\prime} \times 1 \times T^{\prime} \) & \( f^{\prime} \times 2 \times T^{\prime} \) & --- \\ \midrule
            \( \psi_{\mathrm{clf}} \) & 2D convolution & \( (2, 1) \) & \( (1, 1) \) & \( \left( K + 1 \right) \! N_d \) & $f^{\prime} \times 2 \times T^{\prime}$ & $\left( K + 1 \right) \! N_d \times 1 \times T^{\prime}$ & Softmax over \( K + 1 \) \\
            \( \psi_{\mathrm{loc}} \) & 2D convolution & \( (2, 1) \) & \( (1, 1) \) & \( 2 N_d \) & $f^{\prime} \times 2 \times T^{\prime}$ & $2 N_d \times 1 \times T^{\prime}$ & Linear \\ \midrule
            \(\mathbf{z}\) & Output, \(\mathbf{p}\) & --- & --- & --- & $\left( K + 1 \right) \! N_d \times 1 \times T^{\prime}$ & $N_d \times T^{\prime} \times \left( K + 1 \right)$ & --- \\
             & Output, \(\mathbf{y}\) & --- & --- & --- & $2 N_d \times 1 \times T^{\prime}$ & $N_d \times T^{\prime} \times 2$ & --- \\
            \bottomrule
        \end{tabular}
        \begin{tablenotes}
            \small
            \item \(\mathbf{x}\), input containing PSG data; $\mathbf{z}$, output containing predicted arousal probabilities and associated start and duration predictions; \( \phi_{\mathrm{mix}} \), non-linear mixing block; \(\varphi_{\mathrm{conv}}\), convolutional feature extraction block; \( \varphi_{\mathrm{rec}} \) recurrent feature extraction block; \( \psi_{\mathrm{clf}} \), event classification block; \( \psi_{\mathrm{loc}} \), event localization block; $C$, number of input channels; $T$, number of samples in a segment of PSG data; $c$, temporal kernel size; $s$, temporal stride; $f_0$, base number of feature maps; $f^{\prime}=f_02^{k_{\mathrm{max}}}$, maximum number of feature maps; $T^{\prime} = T/s^{k_{\mathrm{max}}}$, reduced temporal dimension in samples; $N_d$, number of default event windows in segment; $K$, number of classes; ReLU, rectified linear unit; bGRU, bidirectional gated recurrent unit.
        \end{tablenotes}
    \end{threeparttable}
\end{table*}
\paragraph*{Notation} We denote by \(\llbracket a, b \rrbracket\) the set of integers \(\lbrace n \in \N \mid a \leq n \leq b\rbrace\) with \(\llbracket N \rrbracket\) being shorthand for \(\llbracket 1, N \rrbracket\), and by \(n \in \llbracket N \rrbracket \) the \(n\)th sample in \(\llbracket N \rrbracket\).
A model for a given experiment is denoted by $\mathcal{M}_{(\cdot)}$, while an optimized model is superscripted with a star as $\mathcal{M}_{(\cdot)}^{*}$.
A segment of PSG data is denoted by $\mathbf{x} \in \real^{C \times T}$, where $C, T$ is the number of channels and the duration of the segment in samples, respectively.

\subsection{Data}
We collected PSGs from \num{1500} subjects in the MrOS Sleep Study~\cite{Blank2005, Orwoll2005, Blackwell2011} from the National Sleep Research Resource repository~\cite{Dean2016, Zhang2018}.
From each PSG, we extracted left and right EEG, left and right EOG, and chin EMG.
EEG and EOG channels were referenced to the contralateral mastoid process.
For each PSG, we also extracted time-stamped arousal scorings containing starts and durations of scored arousal events.
We did not exclude any PSGs from this study based on sleep duration, number of arousal events, or similar criteria.

\subsection{Data partitioning}
The 1500 PSGs were initially partitioned into three subsets \train\textsubscript{1}, \eval\textsubscript{1}, and \test\textsubscript{1} containing 400, 100 and 1000 PSGs, respectively.
Furthermore, we additionally partitioned \test\textsubscript{1} into three smaller subsets \train\textsubscript{2}, \eval\textsubscript{2}, and \test\textsubscript{2} containing 400, 100, and 500 PSGs, respectively.

\subsection{Preprocessing pipeline}
All signals were resampled to \SI{128}{\hertz} using poly-phase filtering with a Kaiser window ($\beta = 5.0$) prior to subsequent processing.
Extracted EEG and EOG signals were filtered with \nth{2} order Butterworth IIR bandpass filters with cutoff frequencies \SI{0.3}{\hertz} and \SI{35}{\hertz}. 
Chin EMG was filtered with a \nth{4} order Butterworth IIR highpass filter with a cutoff frequency of \SI{10}{\hertz}.
Filtered signals were subsequently standardized by 
\begin{equation}
    \mathbf{x}^{(i)} = \frac{\tilde{\mathbf{x}}^{(i)} - \boldsymbol{\mu}^{(i)}}{\boldsymbol{\sigma}^{(i)}},
\end{equation}
where $\tilde{\mathbf{x}}^{(i)} \in \real^{C \times T}$ is the raw matrix containing $C$ input channels and $T$ samples, and $\boldsymbol{\mu}^{(i)}, \boldsymbol{\sigma}^{(i)} \in \real^{C}$ are the mean and standard deviation vectors for the \textit{i}'th PSG, respectively.

\subsection{Model setup}

We expand upon previous work using similar models for sleep event detection~\cite{Chambon2018b, Chambon2019, Olesen2019}. 
Briefly, the model takes as input a tensor of PSG data $\mathbf{x} \in \real^{C \times T}$ and outputs 
\begin{equation}
    \mathbf{z} = \left( \mathbf{p}, \mathbf{y} \right) \in \real^{N_d \times T^{\prime} \times (K+1)} \times \real^{N_d \times T^{\prime} \times 2}
\end{equation}
containing predicted arousal probabilities $\mathbf{p}$ and associated start and durations for predicted arousal events $\mathbf{y}$.
The differentiable function underlying the model comprises a deep neural network architecture consisting of the following modules:
\paragraph{Input mixing module}
Here, non-linear combinations of the input PSG data $\mathbf{x}$ are made using a non-linear mixing block $\phi_{\mathrm{mix}} : \real^{1 \times C \times T} \to \real^{C \times 1 \times T}$.
\paragraph{Feature extraction module}
This module contains two components. 
The first is a convolutional feature extraction block $\varphi_{\mathrm{conv}} : \real^{C \times 1 \times T} \to \real^{f^{\prime} \times 1 \times T^{\prime}}$ consisting of $k_{\mathrm{max}}$ successions of convolutional, batch normalization, and rectified linear unit (ReLU) layers.
Second is a recurrent feature extraction block $\varphi_{\mathrm{rec}} : \real^{f^{\prime} \times 1 \times T^{\prime}} \to \real^{f^{\prime} \times 2 \times T^{\prime}}$ with $f^{\prime}=f_02^{k_{\mathrm{max}}}$ hidden units.
The $\varphi_{\mathrm{conv}}$ block is responsible for bulk feature extraction and temporal decimation using strided convolutions, while $\varphi_{\mathrm{rec}}$ processes the raw features across the reduced temporal dimension using a bidirectional gated recurrent unit~\cite{Cho2014} with $f^{\prime}$ hidden units.
\paragraph{Event detection module}
The output from $\varphi_{\mathrm{rec}}$ is processed by two separate blocks: $ \psi_{\mathrm{clf}} : \real^{f^{\prime} \times 2 \times T^{\prime}} \to \real^{\left( K + 1 \right) \! N_d \times 1 \times T^{\prime}} $ outputs the tensor $\mathbf{p}$ containing predicted arousal probabilities for each time point $t \in \llbracket T^{\prime} \rrbracket$ for each default event window. 
$ \psi_{\mathrm{loc}} : \real^{f^{\prime} \times 2 \times T^{\prime}} \to \real^{2N_d \times T^{\prime}} $ outputs the tensor $\mathbf{y}$ containing predicted start time and durations of arousal events. 
Both $\psi_{\mathrm{clf}}$ and $\psi_{\mathrm{loc}}$ are implemented using $\left(2, 1\right)$ convolutions rather than convolutions over the entire volume as in~\cite{Chambon2018b, Chambon2019, Olesen2019}.
This serves a dual purpose: the first is to reduce the number of parameters to make the network more memory-efficient, while the second purpose is to allow the kernel and feature maps to be temporally invariant.

For a detailed description of the network architecture, see~\cref{tab:architecture}.

\subsection{Loss objective}
The network parameters were optimized according to a three-component loss objective comprising a localization loss \( \ell_{\mathrm{loc}} \) and a positive and negative classification loss $\ell_{+}$ and $\ell_{-}$, respectively, such that
\begin{equation}
    \ell = \ell_{\mathrm{loc}} + \ell_{+} + \ell_{-}.\label{eq:loss}
\end{equation}
The localization loss was calculated using a Huber function
\begin{equation}
    \ell_{\mathrm{loc}} = \frac{1}{N_+} \sum_{i \in \pi_+}\!{h^{(i)}} \\
\end{equation} 
\begin{equation}
    \mathbf{h} =
    \begin{cases}
        0.5 \! \left( \mathbf{y} - \mathbf{t} \right)^2, & \text{if } \lvert \mathbf{y} - \mathbf{t} \rvert < 1, \\
        \lvert \mathbf{y} - \mathbf{t} \rvert - 0.5, & \text{otherwise,}
    \end{cases}
\end{equation}
where $i \in \pi_+$ indicates event windows with a non-empty arousal target.
Contributions from the positive/negative classification losses were calculated using a focal loss function~\cite{Lin2020}:
\begin{align}
    \ell_{+} &= \frac{1}{N_+} \sum_{i \in \pi_+}\!{-\alpha \parentheses*{1 - p^{(i)}}^{\gamma} \func{\log}{p^{(i)}}}, \, \text{and} \\
    \ell_{-} &= \frac{1}{N_{-}} \sum_{i \in \pi_{-}}\!{-\alpha \parentheses*{1 - p^{(i)}}^{\gamma} \func{\log}{p^{(i)}}},
\end{align}
where $\alpha=0.25$ and $\gamma=2$, \(i \in \pi_{-}\) indicates event windows with empty arousal targets, and \( p^{(i)} \in \mathbf{p} \) is the predicted class probabilities for event window \textit{i}.
This serves to counter the class imbalance in a single data segment, which typically consists of many event windows with few positive examples.

\subsection{Experimental setups}
We investigated the channel mismatch problem with the following four experimental setups:
\paragraph{Full montage baseline (FM)}
In this experiment, we trained the event detection algorithm on $\train_1$ using \(C=5\) channels: left/right central EEG, left/right EOG, and chin EMG.
Convergence and the optimal detection threshold were assessed on $\eval_1$ and performance was evaluated on \test\textsubscript{2}.
The optimal baseline model was used as an initialization for the two transfer learning experiments described below.
\paragraph{Pretraining (PT)}
The optimal model $\mbl^{*}$ was used in this experiment as an initialization for $\mpt$.
We adjusted the mixing module and first convolutional layer in the feature extraction module to account for the channel mismatch by replacing the convolutional and batch normalization layers, and subsequently trained these from scratch.
The rest of the weights and bias terms were frozen to the optimized values from $\mbl^{*}$.
The network was trained on $\train_2$ with only $C=1$ channels (left central EEG, C3).
Convergence and optimal detection threshold were assessed on $\eval_2$, while final performance was evaluated on \test\textsubscript{2}.
\paragraph{Fine-tuning (FT)}
Similar to PT, the optimal model $\mbl^{*}$ was used in this experiment as an initialization for $\mft$.
Also, the mixing module and first convolutional layer in the feature extraction module were likewise adjusted.
However, all other layers in $\mft$ were permitted to be further optimized by fine-tuning weights and bias terms during training.
The model was trained using the same 400 PSGs from $\train_2$ with the same $C=1$ channel configuration as in PT.
\paragraph{Single EEG benchmark (SE)}
We benchmarked our two transfer learning experiments to a comparable situation in which an event detection model was trained on the same PSGs in $\train_2$ using only the left central EEG (C3).

In all experimental runs, we optimized the loss objective in~\cref{eq:loss} using the Adam optimization algorithm with a learning rate of \(\alpha=10^{-3}\) and the default parameter values \( \left( \beta_1, \beta_2 \right) = \left( 0.9, 0.999 \right) \) as suggested in~\cite{Kingma2015}.
We applied the same data sampling strategy as proposed in~\cite{Olesen2019}, in which a segment of data is sampled such that it contains at least 50\% of a randomly sampled event across all PSGs.
We used a default event window size of \SI{15}{\second} with \SI{50}{\percent} overlap as this was found previously to work well for arousal detection~\cite{Olesen2019}.
Convergence was assessed for all experimental runs using the loss on the corresponding validation datasets.

All experiments were implemented in PyTorch 1.2~\cite{Paszke2019}.

\subsection{Performance evaluation}
Bipartite matching were used to match detected and true events during training and testing.
At test time, detected events were subjected to non-maximum suppression based on an intersection-over-union (IOU) of at least 0.5 between detected and true events.
We evaluated the performance of our experimental setups using precision, recall and F1 scores.

\subsection{Statistical analysis}
We used Kruskal–Wallis one-way analysis of variance tests for differences in performance metrics between groups (SE, FT and PT) with a significance level of $\alpha=0.05$.
Post-hoc testing was performed with Mann-Whitney U-tests for each pair-combination (SE/FT, SE/PT, and FT/PT) likewise with $\alpha=0.05$.
We accounted for multiple comparisons by adjusting \textit{p}-values with Bonferroni corrections.

\section{RESULTS AND DISCUSSION}

\begin{figure}[tb]
    \centering
    \includegraphics[width=\columnwidth]{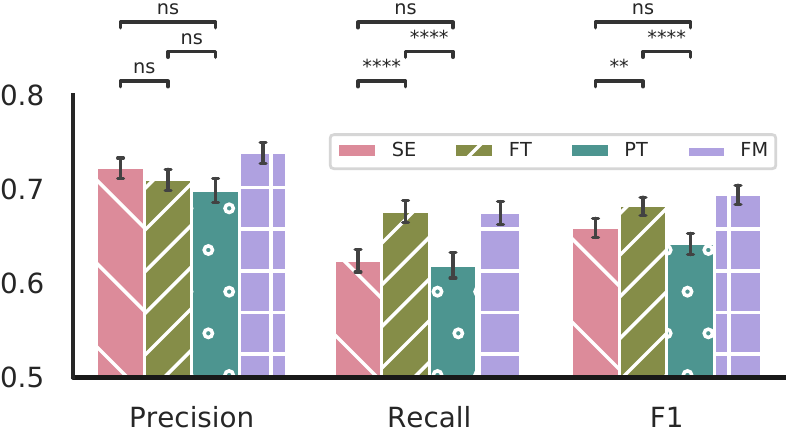}
    \caption{Performance metrics as evaluated on \test\textsubscript{2} for each experimental setup. Bars indicate mean performance across PSGs with 95\% confidence interval error bars. Note the $y$-axis scaling. SE: single-EEG. FT: fine-tuning. PT: pre-training. FM: full montage. ns: not significant, $^{**}$: $p_{\mathrm{adj}} < 10^{-2}$; $^{****}$: $p_{\mathrm{adj}} < 10^{-4}$.}
    \label{fig:results}
\end{figure}

\begin{table}[tb]
    \begin{threeparttable}
        \centering
        \footnotesize
        \caption{Performance metrics across experiments.}
        \label{tab:results}
        \begin{tabular}{@{}lccc@{}}
        \toprule
        \textbf{Experiment} &        \textbf{Precision} &           \textbf{Recall} &               \textbf{F1} \\ \midrule
        FM &  $0.739 \pm 0.122$ &  $0.675 \pm 0.139$ &  $0.694 \pm 0.115$ \\
        SE &  $0.723 \pm 0.124$ &  $0.624 \pm 0.137$ &  $0.659 \pm 0.117$ \\ \midrule
        FT &  $\mathbf{0.710 \pm 0.128}$ &  $\mathbf{0.676 \pm 0.130}$ & $\mathbf{0.682 \pm 0.110}$ \\
        PT &  $0.699 \pm 0.141$ &  $0.619 \pm 0.153$ &  $0.642 \pm 0.129$ \\
        \bottomrule
        \end{tabular}
        \begin{tablenotes}
            \small
            \item Metrics are shown as mean $\pm$ standard deviation when evaluated on each PSG in \test\textsubscript{2}. Best performing transfer learning experiment is shown in bold. SE: single-EEG. FT: fine-tuning. PT: pre-training. FM: full montage.
        \end{tablenotes}
    \end{threeparttable}
\end{table}

We show the results of the transfer learning experiments (FT, PT) as well as the baseline and benchmark experiments (FM, SE) in~\cref{fig:results} and~\cref{tab:results}. 
Performance metrics were not calculated for \num{10} subjects in \test\textsubscript{2}, as these did not have any scored arousals and are thus not reflected in~\cref{fig:results} and~\cref{tab:results}.

The baseline F1 performance is shown to be slightly lower than previously reported ($0.694 \pm 0.115$ vs. $0.749 \pm 0.105$)~\cite{Olesen2019}.
However, our baseline model was trained on \num{400} subjects compared to \num{1485} in~\cite{Olesen2019}, which would account for the lower F1 score. 
By reducing the available input channels from $C=5$ different modalities to $C=1$ EEG channels as in the SE benchmark experiment, the F1 score drops to $0.659 \pm 0.117$, while the precision and recall scores likewise drop from $0.739 \pm 0.122$ to $0.723 \pm 0.124$, and $0.675 \pm 0.139$ to $0.624 \pm 0.137$, respectively.

We found statistically significant differences in F1 scores between SE, FT, and PT ($p=3.189\times 10^{-7}$). 
Post-hoc testing further revealed statistically significant differences between SE and FT ($p_{\mathrm{adj}}=2.224 \times 10^{-3}$), and FT and PT ($p_{\mathrm{adj}}=2.685 \times 10^{-7}$), but not between SE and PT  ($p_{\mathrm{adj}}=0.080$). 
We also found that recall scores differed between experimental setups ($p=7.085 \times 10^{-13}$).
Post-hoc testing showed statistically significant differences between SE, FT ($p_{\mathrm{adj}}=5.180 \times 10^{-11}$), and FT and PT ($p_{\mathrm{adj}}=1.440 \times 10^{-9}$), but not between SE and PT ($p_{\mathrm{adj}}=1.000$).
Lastly, we saw statistically significant differences in precision scores between experimental setups ($p=0.033$), subsequent post-hoc testing did not reveal any statistical significant differences, when adjusting for multiple comparisons using the Bonferroni procedure (SE/FT, $p_{\mathrm{adj}}=0.214$; FT/PT, $p_{\mathrm{adj}}=1.000$; SE/PT, $p_{\mathrm{adj}}=0.037$).

Our results show, that for some scenarios, we can learn information present in multi-variate PSG data and effectively transfer that information to a target domain containing only a single EEG channel.
Specifically, the performance of our fine-tuning strategy is high enough that the mean F1 scores across subjects are statistically insignificant, when comparing FT and FM setups (not shown).

Previous related work focused on the channel mismatch problem, when comparing different, but the same number of, channel modalities such as transferring EEG-based models to EOG-based target domains, and thus did not investigate how changing the model architecture might impact performance~\cite{Phan2019, Phan2019c}.
In this work, we investigated transfer learning when the source and target domains only overlap by one input channel.
This necessitates changing some parts of the underlying model architecture to accommodate the different number of input channels, and these changes might impact downstream feature extraction.
We did not explore simply zeroing out a large number of input channels in this work, as this requires exhaustive search of which channel indices to zero out in the model based on the number of target input channels. 
Our strategy does not require this exhaustive search.

Our study applied a simple optimization strategy for the transfer learning experiments, which might limit the potential performance gain.
This is especially relevant for the FT experiment. 
For example, one could experiment with different learning rates and scheduling schemes for the initial layers and pre-trained layers, such that the initial layers were trained with a higher relative learning rate to compensate for their lack of initial training.

Furthermore, we explored transfer learning for the channel mismatch problem in a single cohort of patient recordings. 
Future directions of this research will investigate scenarios, where both the source and target domains, and the datasets are different.

\section{CONCLUSIONS}
We show in our experiments that a simple fine-tuning strategy can be employed to transfer learning from a model based on multi-variate PSG data to a configuration where only a single EEG lead is available for detecting arousals, and that the difference between single-EEG and multivariate PSG performance is negligible.
Future work will explore the effects of various combinations of datasets on the impact of generalized event detection, when the source and target domains do not overlap completely.

\addtolength{\textheight}{-5.5cm}   



\section*{ACKNOWLEDGMENTS}

Some of the computing for this project was performed on the Sherlock cluster.
We would like to thank Stanford University and the Stanford Research Computing Center for providing computational resources and support that contributed to these research results.

We would also like to thank the National Sleep Research Resource team for their efforts in collecting, organizing and making available the PSG data used in this study.

\section*{REFERENCES}
\printbibliography[heading=none]

\end{document}